\documentclass[journal]{IEEEtran}
\usepackage{subcaption,booktabs}
\usepackage{siunitx} 
\usepackage{amsmath,amsfonts}
\usepackage{amssymb}
\usepackage{algorithmic}
\usepackage{algorithm}
\usepackage{array}
\usepackage{textcomp}
\usepackage{stfloats}
\usepackage{url}
\usepackage{verbatim}
\usepackage{graphicx}
\usepackage{cite}
\usepackage{multirow}
\usepackage{enumitem}
\usepackage{xcolor}
\usepackage{tabularx} 
\usepackage{float}
\usepackage[a4paper, top=16.05mm, bottom=28.4mm, left=18.875mm, right=18.875mm]{geometry}

\usepackage{hyperref}
\definecolor{custompink}{RGB}{239,118,186} 
\hypersetup{
    colorlinks=true, 
    linkcolor=blue,  
    citecolor=green, 
    filecolor=magenta, 
    urlcolor=custompink     
}

\title{Learning-based Detection of GPS Spoofing Attack for Quadrotors}

\author{Pengyu Wang$^{1,2}$, Zhaohua Yang$^{1}$, Jialu Li$^{1}$ and Ling Shi$^{1}$, \textit{Fellow, IEEE}
\thanks{$^{1}$Pengyu Wang, Zhaohua Yang, Jialu Li and Ling Shi are with the Department of Electronic and Computer Engineering, Hong Kong University of Science and Technology, Hong Kong SAR. {\tt\small \{pwangat, zyangcr, jlikr\}@connect.ust.hk, eesling@ust.hk}}
\thanks{$^{2}$Pengyu Wang is also with the Shenzhen Key Laboratory of Robotics Perception and Intelligence and the Department of Electronic and Electrical Engineering, Southern University of Science and Technology, Shenzhen, China.}
}

\begin{document}

\maketitle
\thispagestyle{empty}
\pagestyle{empty}

\begin{abstract}
Safety-critical cyber-physical systems (CPS), such as quadrotor UAVs, are particularly prone to cyber attacks, which can result in significant consequences if not detected promptly and accurately. During outdoor operations, the nonlinear dynamics of UAV systems, combined with non-Gaussian noise, pose challenges to the effectiveness of conventional statistical and machine learning methods. To overcome these limitations, we present QUADFormer, an advanced attack detection framework for quadrotor UAVs leveraging a transformer-based architecture. This framework features a residue generator that produces sequences sensitive to anomalies, which are then analyzed by the transformer to capture statistical patterns for detection and classification. Furthermore, an alert mechanism ensures UAVs can operate safely even when under attack. Extensive simulations and experimental evaluations highlight that QUADFormer outperforms existing state-of-the-art techniques in detection accuracy. (Video\footnote{\href{https://youtu.be/SiXHATM4eZQ}{https://youtu.be/SiXHATM4eZQ}})

\end{abstract}

\section{Introduction}
\label{sec:introduction}

\begin{figure}[t!]
\centering
\includegraphics[width=0.98\columnwidth]{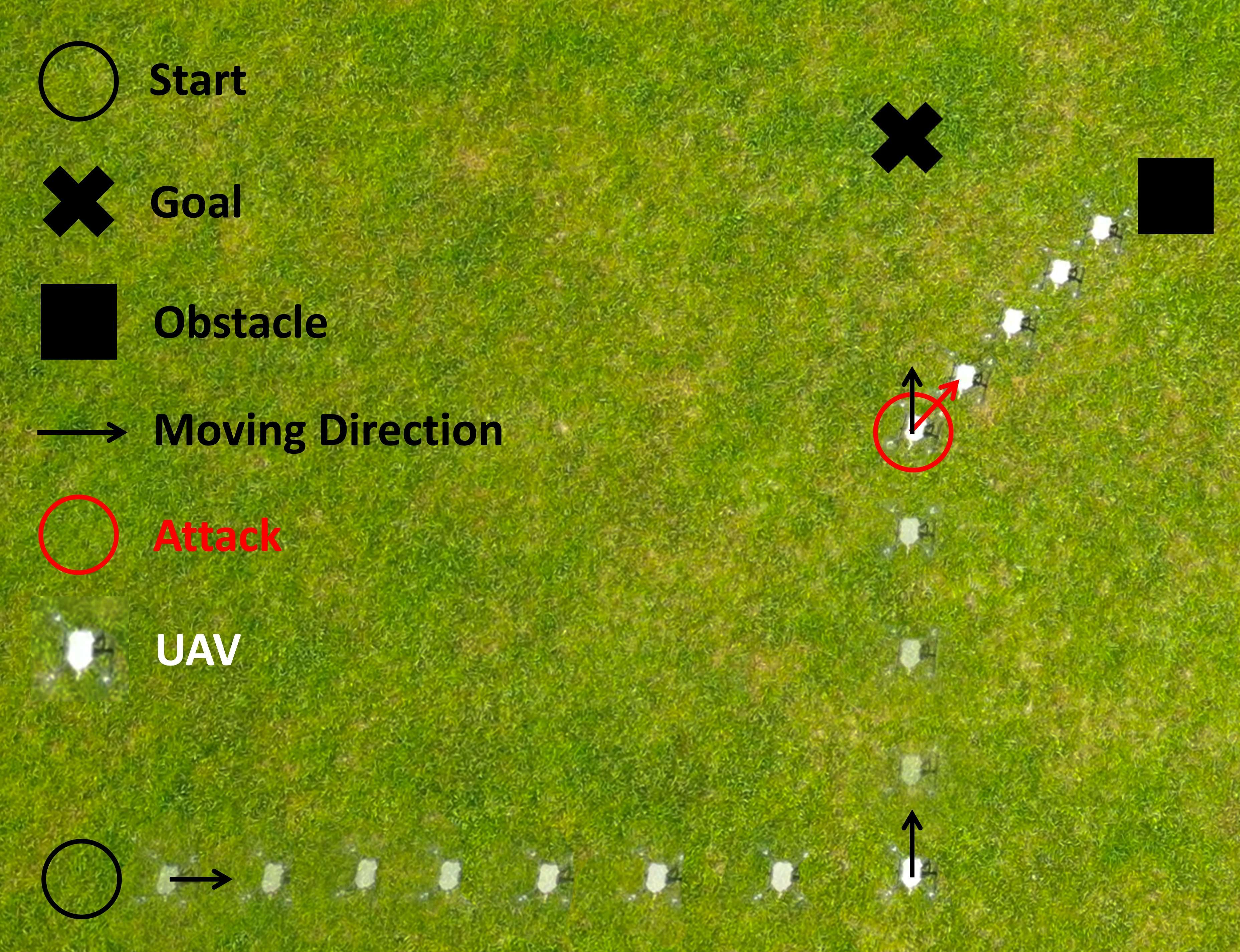} 
\caption{UAV operation under GPS spoofing attacks.}
\label{fig:attack_overview}
\end{figure}

The maneuverability of UAVs has made them increasingly prevalent across diverse applications~\cite{wang2022quadrotor, wang2024miner}. However, for UAVs deployed in safety-critical environments, their inherent vulnerabilities may lead to catastrophic incidents. As illustrated in Fig.\ref{fig:attack_overview}, the openness of GPS signals renders UAVs especially vulnerable to cyber attacks, which have already resulted in substantial damages and even tragic outcomes\cite{hassija2021fast}. Cyber attacks targeting UAVs are typically classified into categories such as denial-of-service (DoS) attacks, false data injection (FDI) attacks, and replay attacks~\cite{griffioen2019tutorial}. Among these, FDI attacks stand out due to their higher prevalence and ease of execution compared to replay attacks, as well as their more covert and deceptive nature relative to DoS attacks. Consequently, our research focuses on addressing these FDI attacks.

To safeguard CPS from malicious attacks, two primary detection approaches have been established: residue-based and learning-based methods. Residue-based techniques, such as Chi-squared tests and CUSUM detectors, rely on analyzing filter residues and triggering alarms when predefined thresholds are breached~\cite{murguia2016characterization, muniraj2017framework, wan2020safety}. However, these methods face significant drawbacks: their assumption of low-dimensional or Gaussian data diminishes their effectiveness in UAV systems with high-dimensional, non-Gaussian dynamics, and they require frequent recalibration of thresholds in dynamic environments, which is impractical for UAV operations. On the other hand, learning-based methods leverage machine learning to identify anomalies~\cite{farivar2019artificial, khoei2022comparative, wu2023highly}, but they encounter issues such as susceptibility to noisy sensor data and reliance on large datasets. Moreover, many existing learning-based approaches emphasize either localized point representations or recursive state transitions, which restrict their capacity for parallel processing and limit contextual awareness. These limitations underscore the necessity of developing a method that integrates the advantages of both approaches while overcoming their inherent shortcomings.

\section{Methodology}

\subsection{Problem Formulation}
The quadrotor UAV system is formulated as follows:
\begin{equation}\label{eq:system_model}
\begin{aligned}
    & x_{k+1} = f(x_k,u_k) + w_k, \\
    & y_k = g(x_k) + v_k,
\end{aligned}
\end{equation}
where \( x_k \in \mathbb{R}^n \) is the state, \( u_k \in \mathbb{R}^p \) is the control input, \( y_k \in \mathbb{R}^m \) is the measurement, all at time step \( k \), $f(\cdot)$ and $g(\cdot)$ are non-linear transition functions of the state model and measurement model, \( w_k \in \mathbb{R}^n \) and \( v_k \in \mathbb{R}^m \) are process noise and measurement noise with covariance \( Q_k \in \mathbb{R}^{n \times n} \succeq 0 \) and \( R_k \in \mathbb{R}^{m \times m} \succ 0 \). 

The attacker manipulates the sensor readings, which could involve GPS spoofing. The measurement equation under the FDI attack is:
\begin{equation}\label{eq:fdi}
    y_k = g(x_k)+ v_k + d_k,
\end{equation}
where \( d_k \in \mathbb{R}^m \) represents the attack vector, being a sparse and persistent false data injection. 

For attack detectors, the typical accuracy indicator is defined as follows: 
\begin{equation}\label{eq:F1}
    \text{F1-score} = \frac{2}{\mathrm{Precision}^{-1} + \mathrm{Recall}^{-1}}.
\end{equation}

Consider system (\ref{eq:system_model}) with possible attacks (\ref{eq:fdi}) acting on the system, design an attack detector such that the performance (\ref{eq:F1}) is maximized.  

\subsection{QUADFormer Framework}

\begin{figure*}[ht!]
\centering
\includegraphics[width=\textwidth]{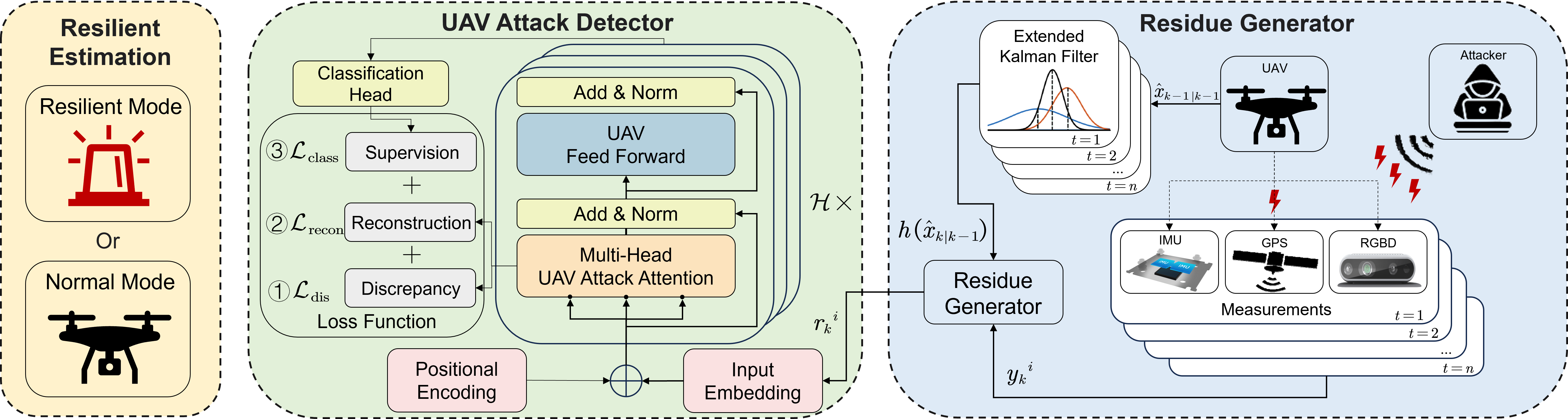}
\caption{Overview of QUADFormer framework.}
\label{fig:frame}
\end{figure*}

Our proposed framework comprises three key components: a residue generator, a UAV attack detector, and a resilient state estimation module, as depicted in Fig.~\ref{fig:frame}. First, an Extended Kalman Filter (EKF) is employed to integrate data from multiple sensors, allowing the UAV to estimate system states and produce meaningful residue sequences. Second, we introduce a semi-supervised transformer-based architecture that incorporates a tailored self-attention mechanism and specifically designed loss functions to perform attack classification and detection. Finally, the resilient state estimation module mitigates the impact of compromised sensors by deactivating them, utilizing secure sensors for pose estimation, and reactivating the compromised sensors once their security is restored.

\subsubsection{Residue Generator}\label{sec:residue_generator}

Compared to raw sensor data, EKF-generated residues possess two critical features that make them particularly suitable for UAV attack detection. First, the EKF's model prediction relies on first-order linearization, leaving higher-order terms in the residue. These terms effectively capture the nonlinear dynamics inherent to UAV systems. Second, since the EKF is derived from the Kalman Filter (KF) and assumes Gaussian noise, under single-mode non-Gaussian noise, the residue exhibits approximate Gaussian properties while retaining the noise approximation error. These attributes provide EKF residues with robust statistical characteristics, making them ideal for extracting features of UAV nonlinear dynamics and non-Gaussian noise. This, in turn, enhances the effectiveness of neural network-based detection methods.

\subsubsection{Attack Detector}\label{sec:attack_detector} 

Our method harnesses the advantages of the transformer architecture for handling sequential data, tailoring it to the unique properties of UAVs and their data. Building upon transformer frameworks designed for anomaly detection, our approach incorporates several key innovations. First, we design a custom attention module to capture the statistical features of attacked points within UAV residue data more effectively. Second, we introduce a novel loss function to facilitate efficient semi-supervised learning. Finally, we propose new criteria for accurately identifying the onset of an attack.

\subsubsection{Resilient State Estimation}\label{sec:state_estimation}

To mitigate the impact of sensor attacks, we incorporate a resilient state estimation system into the UAV, striking a balance between vigilance and responsive efficiency. Under normal operation, this system combines data from GPS, a stereo camera, and an IMU for accurate pose estimation. While visual-inertial odometry provides reliable performance, GPS data remains essential for correcting drift during extended flights. Utilizing GPS alongside a stereo camera and an IMU represents a practical and cost-effective configuration for outdoor UAV operations. In the event of a detected GPS attack, the system excludes the compromised GPS data and relies on the IMU and camera, which are more secure due to their internal placement and reduced susceptibility to external interference.

\section{Results and Analysis}

We replicate the most widely used attack detection approaches, including traditional methods and machine learning-based techniques, to ensure comprehensive evaluation. For traditional state-of-the-art methods, we utilize CUSUM~\cite{yoon2019towards}\cite{liu2019secure}, SPRT~\cite{muniraj2017framework}\cite{kwon2016real}, and BHT~\cite{muniraj2017framework}. For learning-based methods, we replace their original inputs with the residue sequence generated by our residue generator, ensuring fair and consistent comparisons. Specifically, SVM~\cite{khoei2022comparative}, leveraging kernel techniques, offers nonlinear high-dimensional classification capabilities, while CNN~\cite{wu2023highly} and LSTM~\cite{wu2023highly}\cite{wang2020intelligent} effectively extract local features and short-term dependencies. Experimental results demonstrate that our approach achieves exceptional performance across various models, noise levels, and attack intensities, surpassing traditional and learning-based methods in most metrics. Detailed experimental setups and results are provided in the appendix.

\section{Conclusion}

In this paper, we propose a novel framework for cyber attack detection in quadrotor UAVs, combining residue-based and learning-based approaches to address the limitations of each method when used independently. Our framework effectively models the statistical features of UAVs' nonlinear dynamics and non-Gaussian noise, demonstrating superior performance in detecting sensor attacks compared to current state-of-the-art techniques. Future work will involve extending the framework to handle a broader range of cyber attacks and investigating advanced residue generation methods for application in more complex systems.

\section*{Appendix}\label{sec:appendix}

\subsubsection{Real-world Experiment Setup}

\begin{figure}[h]
    \centering
    \hfill 
    \begin{subfigure}[b]{0.49\columnwidth}
        \includegraphics[width=\linewidth]{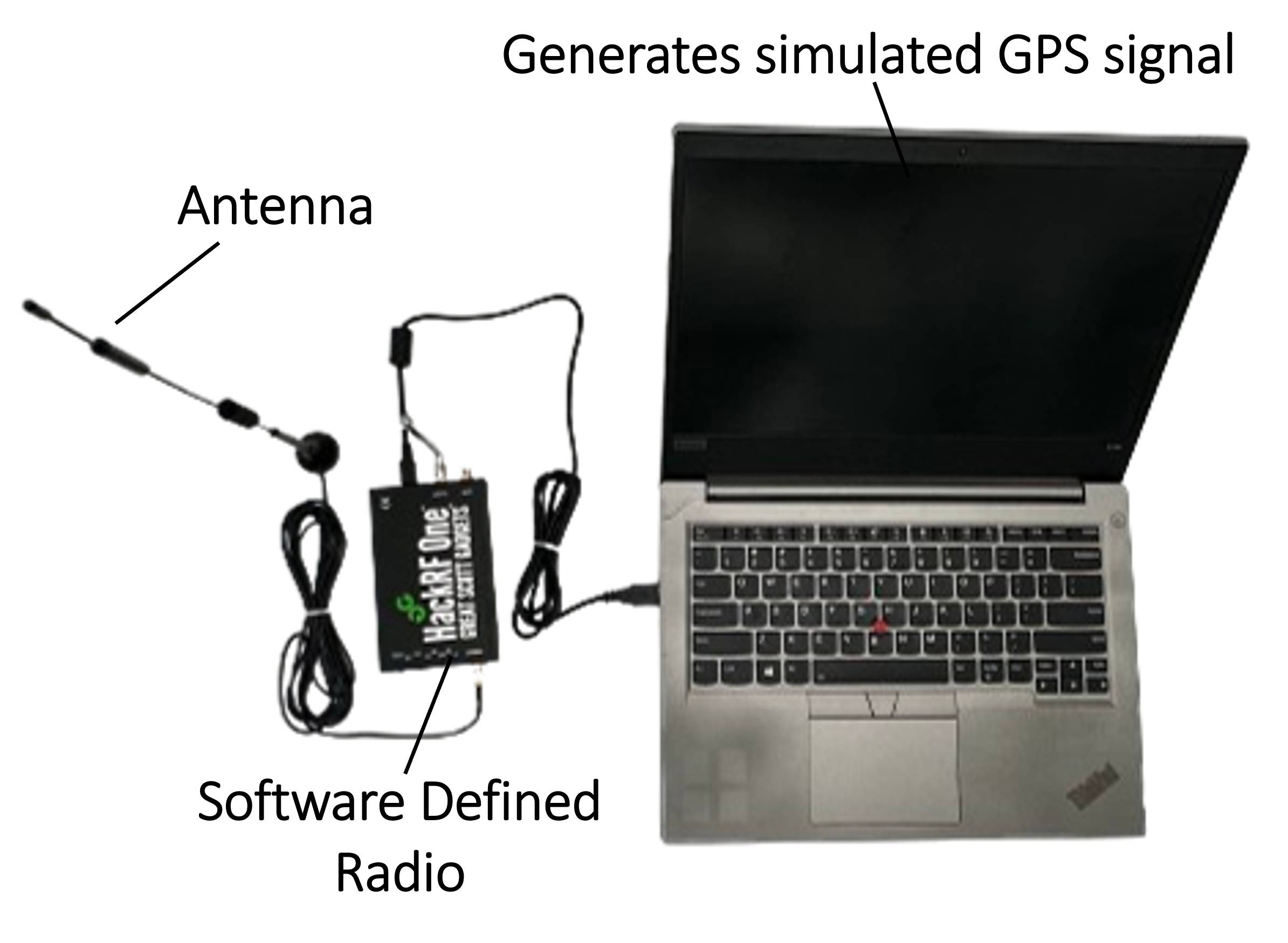}
        \caption{Real attacks launcher.}
        \label{fig:attack_signal}
    \end{subfigure}
    \hfill
    \begin{subfigure}[b]{0.49\columnwidth}
        \includegraphics[width=\linewidth]{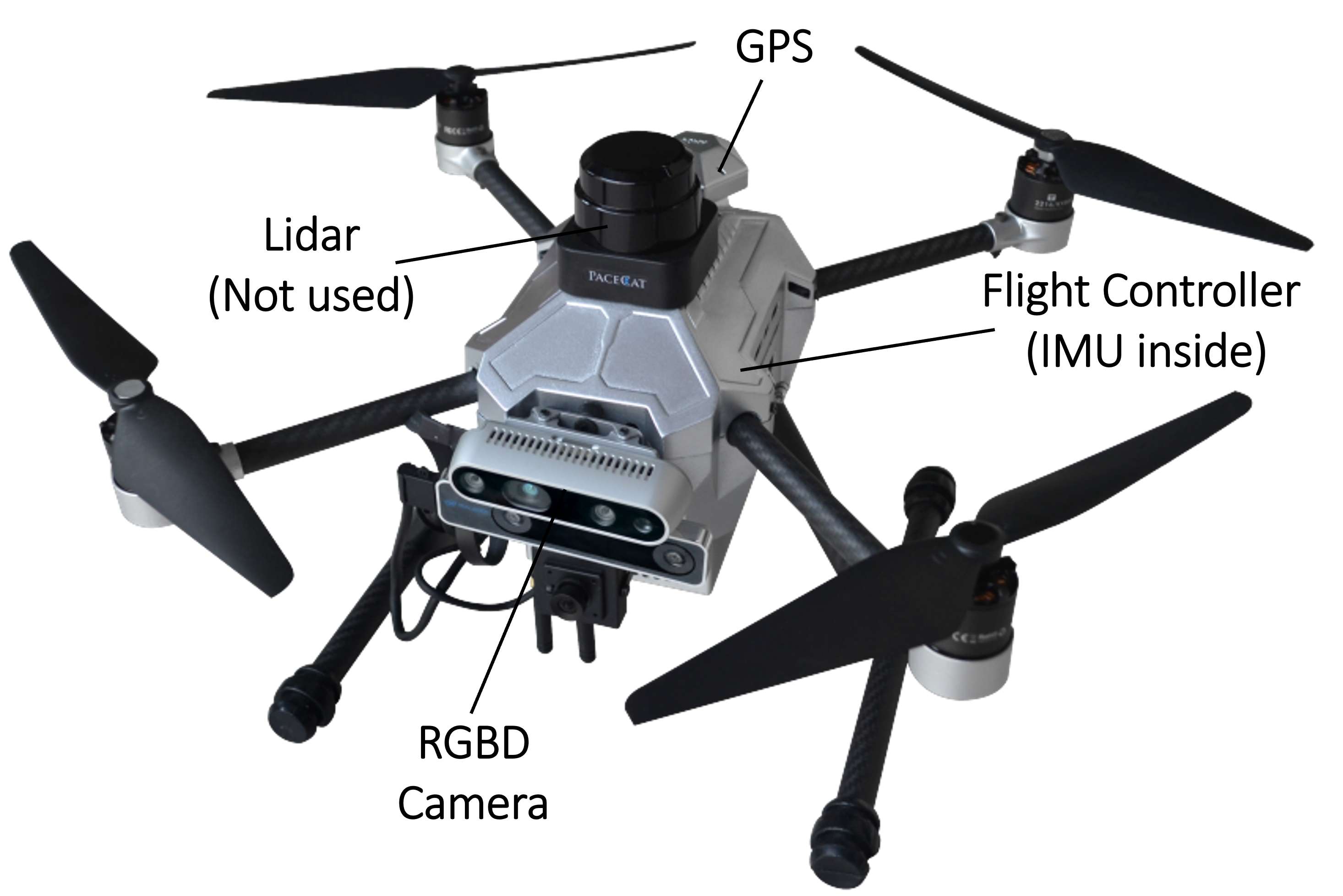}
        \caption{Quadrotor UAV system.}
        \label{fig:uav}
    \end{subfigure}
    \caption{Real-world experiment setup.}
    \label{fig:hardware_combined}
\end{figure}

As shown in Fig.~\ref{fig:hardware_combined}.

\subsubsection{Comparative Result}
As shown in Tab.~\ref{tab:comprehensive_analysis1} - Tab.~\ref{tab:comprehensive_analysis2}.

\begin{table*}[t]
\centering
\caption{Comparison Results for Model I.}
\label{tab:comprehensive_analysis1}

\bgroup
\renewcommand{\arraystretch}{1.5} 

\begin{tabularx}{\textwidth}{|c|c|>{\centering\arraybackslash}X|>{\centering\arraybackslash}X|>{\centering\arraybackslash}X|>{\centering\arraybackslash}X|>{\centering\arraybackslash}X|>{\centering\arraybackslash}X|>{\centering\arraybackslash}X|>{\centering\arraybackslash}X|>{\centering\arraybackslash}X|>{\centering\arraybackslash}X|>{\centering\arraybackslash}X|>{\centering\arraybackslash}X|}
\hline
\multirow{3}{*}{System Model} & UAV Model & \multicolumn{12}{c|}{Model I} \\
\cline{2-14}
 & Noise Model & \multicolumn{6}{c|}{Exponential Noise} & \multicolumn{6}{c|}{Laplacian Noise} \\
\cline{2-14}
 & Attack Model & \multicolumn{3}{c|}{Attack I} & \multicolumn{3}{c|}{Attack II} & \multicolumn{3}{c|}{Attack I} & \multicolumn{3}{c|}{Attack II} \\
\hline
Performance & Metric & P & R & F1 & P & R & F1 & P & R & F1 & P & R & F1 \\
\hline
\multirow{3}{*}{Traditional Method} & CUSUM\cite{yoon2019towards}\cite{liu2019secure} & 0.69 & 0.64 & 0.66 & 0.71 & 0.59 & 0.64 & 0.70 & 0.62 & 0.66 & 0.71 & 0.58 & 0.64 \\
\cline{2-14}
 & SPRT\cite{muniraj2017framework}~\cite{kwon2016real} & 0.69 & 0.67 & 0.69 & 0.68 & 0.68 & 0.67 & 0.69 & 0.67 & 0.68 & 0.68 & 0.67 & 0.67 \\
\cline{2-14}
 & BHT\cite{muniraj2017framework} & 0.62 & 0.65 & 0.64 & 0.68 & 0.60 & 0.65 & 0.66 & 0.59 & 0.62 & 0.70 & 0.60 & 0.65 \\
\hline
\multirow{3}{*}{\parbox{2cm}{\centering Learning-based\\ Method}} & SVM\cite{khoei2022comparative} & 0.88 & 0.82 & 0.85 & 0.75 & 0.69 & 0.72 & 0.88 & 0.82 & 0.85 & 0.75 & 0.69 & 0.72 \\
\cline{2-14}
 & CNN\cite{wu2023highly} & 0.89 & 0.81 & 0.85 & 0.90 & 0.78 & 0.84 & 0.90 & 0.84 & 0.87 & 0.90 & 0.78 & 0.84 \\
\cline{2-14}
 & LSTM\cite{wu2023highly}\cite{wang2020intelligent} & 0.88 & 0.78 & 0.83 & 0.78 & 0.86 & 0.82 & 0.86 & 0.84 & 0.85 & 0.81 & 0.85 & 0.83 \\
\hline
\textbf{Ours} & \textbf{QUADFormer} & \textbf{0.92} & \textbf{0.93} & \textbf{0.93} & \textbf{0.89} & \textbf{0.95} & \textbf{0.92} & \textbf{0.92} & \textbf{0.94} & \textbf{0.93} & \textbf{0.89} & \textbf{0.95} & \textbf{0.92} \\
\hline
\end{tabularx}

\egroup 
\end{table*}

\begin{table*}[t]
\centering
\caption{Comparison Results for Model II.}
\label{tab:comprehensive_analysis2}

\bgroup
\renewcommand{\arraystretch}{1.5} 

\begin{tabularx}{\textwidth}{|c|c|>{\centering\arraybackslash}X|>{\centering\arraybackslash}X|>{\centering\arraybackslash}X|>{\centering\arraybackslash}X|>{\centering\arraybackslash}X|>{\centering\arraybackslash}X|>{\centering\arraybackslash}X|>{\centering\arraybackslash}X|>{\centering\arraybackslash}X|>{\centering\arraybackslash}X|>{\centering\arraybackslash}X|>{\centering\arraybackslash}X|}
\hline
\multirow{3}{*}{System Model} & UAV Model & \multicolumn{12}{c|}{Model II} \\
\cline{2-14}
 & Noise Model & \multicolumn{6}{c|}{Exponential Noise} & \multicolumn{6}{c|}{Laplacian Noise} \\
\cline{2-14}
 & Attack Model & \multicolumn{3}{c|}{Attack I} & \multicolumn{3}{c|}{Attack II} & \multicolumn{3}{c|}{Attack I} & \multicolumn{3}{c|}{Attack II} \\
\hline
Performance & Metric & P & R & F1 & P & R & F1 & P & R & F1 & P & R & F1 \\
\hline
\multirow{3}{*}{Traditional Method} & CUSUM\cite{yoon2019towards}\cite{liu2019secure} & 0.75 & 0.76 & 0.76 & 0.73 & 0.70 & 0.72 & 0.73 & 0.69 & 0.71 & 0.72 & 0.69 & 0.71 \\
\cline{2-14}
 & SPRT\cite{muniraj2017framework}~\cite{kwon2016real} & 0.75 & 0.77 & 0.76 & 0.74 & 0.77 & 0.74 & 0.74 & 0.73 & 0.73 & 0.73 & 0.70 & 0.71 \\
\cline{2-14}
 & BHT\cite{muniraj2017framework} & 0.70 & 0.72 & 0.72 & 0.75 & 0.71 & 0.73 & 0.75 & 0.70 & 0.72 & 0.69 & 0.72 & 0.71 \\
\hline
\multirow{3}{*}{\parbox{2cm}{\centering Learning-based\\ Method}} & SVM\cite{khoei2022comparative} & 0.94 & 0.76 & 0.84 & 0.94 & 0.61 & 0.74 & 0.94 & 0.75 & 0.83 & 0.93 & 0.61 & 0.74 \\
\cline{2-14}
 & CNN\cite{wu2023highly} & 0.93 & 0.73 & 0.82 & 0.94 & 0.70 & 0.80 & 0.96 & 0.67 & 0.79 & 0.85 & 0.72 & 0.78 \\
\cline{2-14}
 & LSTM\cite{wu2023highly}\cite{wang2020intelligent} & 0.89 & 0.80 & 0.84 & 0.90 & 0.75 & 0.82 & 0.81 & 0.83 & 0.82 & 0.89 & 0.74 & 0.81 \\
\hline
\textbf{Ours} & \textbf{QUADFormer} & \textbf{0.92} & \textbf{0.97} & \textbf{0.94} & \textbf{0.89} & \textbf{0.94} & \textbf{0.91} & \textbf{0.91} & \textbf{0.96} & \textbf{0.94} & \textbf{0.89} & \textbf{0.97} & \textbf{0.93} \\
\hline
\end{tabularx}

\egroup 
\end{table*}

\bibliographystyle{IEEEtran}
\bibliography{main}

\begin{thebibliography}{10}
\providecommand{\url}[1]{#1}
\csname url@samestyle\endcsname
\providecommand{\newblock}{\relax}
\providecommand{\bibinfo}[2]{#2}
\providecommand{\BIBentrySTDinterwordspacing}{\spaceskip=0pt\relax}
\providecommand{\BIBentryALTinterwordstretchfactor}{4}
\providecommand{\BIBentryALTinterwordspacing}{\spaceskip=\fontdimen2\font plus
\BIBentryALTinterwordstretchfactor\fontdimen3\font minus \fontdimen4\font\relax}
\providecommand{\BIBforeignlanguage}[2]{{%
\expandafter\ifx\csname l@#1\endcsname\relax
\typeout{** WARNING: IEEEtran.bst: No hyphenation pattern has been}%
\typeout{** loaded for the language `#1'. Using the pattern for}%
\typeout{** the default language instead.}%
\else
\language=\csname l@#1\endcsname
\fi
#2}}
\providecommand{\BIBdecl}{\relax}
\BIBdecl

\bibitem{wang2022quadrotor}
P.~Wang, C.~Wang, J.~Wang, and M.~Q.-H. Meng, ``Quadrotor autonomous landing on moving platform,'' \emph{Procedia Computer Science}, vol. 209, pp. 40--49, 2022.

\bibitem{wang2024miner}
P.~Wang, J.~Tang, H.~W. Lin, F.~Zhang, C.~Wang, J.~Wang, L.~Shi, and M.~Q.-H. Meng, ``{MINER-RRT*}: A hierarchical and fast trajectory planning framework in 3d cluttered environments,'' \emph{arXiv preprint arXiv:2406.00706}, 2024.

\bibitem{hassija2021fast}
V.~Hassija, V.~Chamola, A.~Agrawal, A.~Goyal, N.~C. Luong, D.~Niyato, F.~R. Yu, and M.~Guizani, ``Fast, reliable, and secure drone communication: A comprehensive survey,'' \emph{IEEE Communications Surveys \& Tutorials}, vol.~23, no.~4, pp. 2802--2832, 2021.

\bibitem{griffioen2019tutorial}
P.~Griffioen, S.~Weerakkody, B.~Sinopoli, O.~Ozel, and Y.~Mo, ``A tutorial on detecting security attacks on cyber-physical systems,'' in \emph{18th European Control Conference}, 2019, pp. 979--984.

\bibitem{murguia2016characterization}
C.~Murguia and J.~Ruths, ``Characterization of a {CUSUM} model-based sensor attack detector,'' in \emph{IEEE 55th Conference on Decision and Control}, 2016, pp. 1303--1309.

\bibitem{muniraj2017framework}
D.~Muniraj and M.~Farhood, ``A framework for detection of sensor attacks on small unmanned aircraft systems,'' in \emph{IEEE International Conference on Unmanned Aircraft Systems}, 2017, pp. 1189--1198.

\bibitem{wan2020safety}
W.~Wan, H.~Kim, N.~Hovakimyan, L.~Sha, and P.~G. Voulgaris, ``A safety constrained control framework for {UAV}s in {GPS} denied environment,'' in \emph{59th IEEE Conference on Decision and Control}, 2020, pp. 214--219.

\bibitem{farivar2019artificial}
F.~Farivar, M.~S. Haghighi, A.~Jolfaei, and M.~Alazab, ``Artificial intelligence for detection, estimation, and compensation of malicious attacks in nonlinear cyber-physical systems and industrial {IoT},'' \emph{IEEE Transactions on Industrial Informatics}, vol.~16, no.~4, pp. 2716--2725, 2019.

\bibitem{khoei2022comparative}
T.~T. Khoei, A.~Gasimova, M.~A. Ahajjam, K.~Al~Shamaileh, V.~Devabhaktuni, and N.~Kaabouch, ``A comparative analysis of supervised and unsupervised models for detecting {GPS} spoofing attack on {UAVs},'' in \emph{IEEE International Conference on Electro Information Technology}, 2022, pp. 279--284.

\bibitem{wu2023highly}
S.~Wu, Y.~Li, Z.~Wang, Z.~Tan, and Q.~Pan, ``A highly interpretable framework for generic low-cost {UAV} attack detection,'' \emph{IEEE Sensors Journal}, vol.~23, no.~7, pp. 7288--7300, 2023.

\bibitem{yoon2019towards}
H.-J. Yoon, W.~Wan, H.~Kim, N.~Hovakimyan, L.~Sha, and P.~G. Voulgaris, ``Towards resilient {UAV}: Escape time in {GPS} denied environment with sensor drift,'' \emph{IFAC-PapersOnLine}, vol.~52, no.~12, pp. 423--428, 2019.

\bibitem{liu2019secure}
Q.~Liu, Y.~Mo, X.~Mo, C.~Lv, E.~Mihankhah, and D.~Wang, ``Secure pose estimation for autonomous vehicles under cyber attacks,'' in \emph{IEEE Intelligent Vehicles Symposium}, 2019, pp. 1583--1588.

\bibitem{kwon2016real}
C.~Kwon, S.~Yantek, and I.~Hwang, ``Real-time safety assessment of unmanned aircraft systems against stealthy cyber attacks,'' \emph{Journal of Aerospace Information Systems}, vol.~13, no.~1, pp. 27--45, 2016.

\bibitem{wang2020intelligent}
S.~Wang, J.~Wang, C.~Su, and X.~Ma, ``Intelligent detection algorithm against {UAVs}' {GPS} spoofing attack,'' in \emph{IEEE 26th International Conference on Parallel and Distributed Systems}, 2020, pp. 382--389.

\end{thebibliography}

\end{document}